\documentclass[12pt,a4paper]{article}
\pdfoutput=1
\usepackage[utf8]{inputenc}
\usepackage[T1]{fontenc}
\usepackage{amsmath,amssymb}
\usepackage{graphicx}
\usepackage{booktabs}
\usepackage{hyperref}
\usepackage{xcolor}
\usepackage{geometry}
\usepackage{natbib}
\usepackage{caption}
\usepackage{array}
\usepackage{url}
\hypersetup{colorlinks=true,linkcolor=blue,citecolor=blue,urlcolor=blue}

\title{Early Rug Pull Warning for BSC Meme Tokens via\\Multi-Granularity Wash-Trading Pattern Profiling}
\author{Dingding Cao\textsuperscript{a}\quad Bianbian Jiao\textsuperscript{b}\quad Jingzong Yang\textsuperscript{a}\\Yujing Zhong\textsuperscript{a}\quad Wei Yang\textsuperscript{a,*}\\[6pt]\textsuperscript{a}School of Big Data, Baoshan University, Baoshan 678000, Yunnan, China\\\textsuperscript{b}School of Information Science and Electrical Engineering,\\Shandong Jiaotong University, Jinan, Shandong, China\\[4pt]\textsuperscript{*}Corresponding author. E-mail: bsxy10202@bsc.edu.cn}
\date{}

\begin{document}
\maketitle

\begin{abstract}
The high-frequency issuance and short-cycle speculation of meme tokens in decentralized finance (DeFi) have significantly amplified rug-pull risk. Existing approaches still struggle to provide stable early warning under scarce anomalies, incomplete labels, and limited interpretability. To address this issue, an end-to-end warning framework is proposed for BSC meme tokens, consisting of four stages: dataset construction and labeling, wash-trading pattern feature modeling, risk prediction, and error analysis. Methodologically, 12 token-level behavioral features are constructed based on three wash-trading patterns (Self, Matched, and Circular), unifying transaction-, address-, and flow-level signals into risk vectors. Supervised models are then employed to output warning scores and alert decisions. Under the current setting (7 tokens, 33,242 records), Random Forest outperforms Logistic Regression on core metrics, achieving $\text{AUC}=0.9098$, $\text{PR-AUC}=0.9185$, and $F_1=0.7429$. Ablation results show that trade-level features are the primary performance driver ($\Delta\text{PR-AUC}=-0.1843$ when removed), while address-level features provide stable complementary gain ($\Delta\text{PR-AUC}=-0.0573$). The model also demonstrates actionable early-warning potential for a subset of samples, with a mean Lead Time~(v1) of 3.8133 hours. The error profile ($\text{FP}=1$, $\text{FN}=8$) indicates that the current system is better positioned as a high-precision screener rather than a high-recall automatic alarm engine. The main contributions are threefold: an executable and reproducible rug-pull warning pipeline, empirical validation of multi-granularity wash-trading features under weak supervision, and deployment-oriented evidence through lead-time and error-bound analysis.
\end{abstract}


\section{Introduction}

Decentralized finance (DeFi) has demonstrated strong innovation potential in openness and composability, yet its low entry barriers and high anonymity also amplify on-chain fraud risks. This problem is particularly severe in the high-frequency issuance and short-cycle speculation of meme tokens, where rug-pull events often spread rapidly, concentrate losses among retail participants, and remain difficult to trace and prosecute~\cite{cernera2023token,mazorra2022rug,xia2021trade}. On low-fee public chains such as BNB Smart Chain (BSC), transaction frictions are further reduced, allowing abnormal fund behaviors to escalate and propagate within a short time. Providing actionable warning signals before risk events occur has therefore become a critical problem for on-chain risk control and regulatory technology.

Existing studies and engineering practices in this area mainly follow three directions. The first is contract-level static analysis for identifying high-risk privileges, suspicious tax mechanisms, and potential backdoors. The second is address-network and transaction-graph modeling for abnormal interaction patterns. The third involves supervised models or rule engines for risk classification and ranking~\cite{luo2025ai,sun2024sok,zhou2023sokdefi}. Although these approaches provide useful foundations for on-chain security analysis, three common challenges remain in the meme-token context: scarce anomalies with incomplete labels, fast risk evolution that weakens static features, and limited interpretability that hinders the mapping of warnings to verifiable behavioral evidence chains.

Research on on-chain fraud detection generally follows a pipeline of suspicious-entity identification, abnormal-behavior characterization, and risk-score output. In rug-pull scenarios, features are typically constructed from historical incidents and then fed into classifiers to identify high-risk tokens or projects~\cite{luo2025ai,nguyen2023rugpull,kalacheva2025detecting}. Such methods are practical and deployment-friendly, but their performance is highly sensitive to label quality and sample coverage. When anomaly samples are scarce or definitions are inconsistent, threshold stability and cross-dataset transferability often degrade. Another line of work emphasizes event-level evidence chains, integrating liquidity withdrawal, privilege changes, and transaction bursts along timelines to improve the trustworthiness and interpretability of alerts~\cite{xia2021trade,srifa2025rugpull,zhou2024stop}. This direction is valuable for risk-control operations but usually requires finer-grained annotations and higher data-collection cost. In fast-cycle meme-token markets, balancing evidence sufficiency and real-time usability remains an open challenge.

Static smart-contract analysis represents a major branch of on-chain risk research, focusing on ownership privileges, upgradeable backdoors, tax logic, and blacklist controls~\cite{ding2023survey,sun2023smart,feist2019slither}. These methods can provide structured early warnings in pre-screening stages, yet their limitations are also clear: static risks do not always materialize into short-term observable incidents, and explainability can be limited for proxy patterns, obfuscated logic, or multi-contract interactions. For rug-pull warning, contract-level signals are often combined with behavioral features to reduce blind spots of single-path analysis~\cite{sun2024sok,qian2023cross,zheng2022securing}. However, many existing fusion strategies remain coarse, causing either over-dominance of contract priors or masking of mechanism risks by behavioral noise. Designing scalable and calibratable multi-source feature fusion remains a key research gap.

Address-network and transaction-graph methods identify abnormal fund flows via node relations, edge distributions, and community structures, and are among the most common technical approaches in on-chain risk control~\cite{wu2021analysis,chen2020understanding,han2024mt2ad}. Graph statistics or graph embeddings are typically constructed for downstream classification and ranking. The strength of these approaches lies in modeling coordinated group behavior and cross-address associations. Nevertheless, practical challenges arise in meme-token contexts. Token lifecycles are short and graph structures evolve rapidly, so static snapshots may miss critical transition points. Graph-model outputs are also less straightforward to translate into actionable operational decisions, whereas lightweight behavioral-feature approaches can offer better real-time maintainability and operational interpretability.

Wash-trading studies commonly categorize manipulative behavior into Self, Matched, and Circular patterns to represent distinct mechanisms~\cite{victor2021detecting,cong2023crypto,cui2022wteye}. The main advantage of this line of research is interpretability, as each pattern can be mapped to intuitive transactional evidence that facilitates human review and audit tracing. A key limitation, however, is that many studies stop at pattern detection and do not fully address whether such pattern features can provide stable early-warning signals. A methodological gap thus remains between detecting wash trading and warning of rug pulls in advance, especially under incomplete labels and rare anomalies. More recent work has extended wash-trading analysis to NFT markets and flash-loan-based manipulation~\cite{tosic2025beyond,vonwachter2022nft,gan2022flash}, yet the bridge from pattern profiling to actionable early-warning scoring has not been systematically validated.

In the broader context of on-chain risk, anomaly detection surveys~\cite{ulhassan2023anomaly} and Ponzi scheme detection studies~\cite{chen2018ponzi,bartoletti2020dissecting} have demonstrated the effectiveness of behavioral-feature-based classification on Ethereum. DeFi ecosystem analyses~\cite{werner2022sok,xu2023sokdex} have further highlighted the structural vulnerabilities of automated market makers and decentralized exchanges that enable rug-pull operations. Frontrunning and miner-extractable-value research~\cite{daian2020flash} has also revealed how transaction-ordering dependencies create exploitable attack surfaces. These studies collectively inform the present work by establishing the importance of transaction-level behavioral signals for on-chain risk modeling.

Overall, prior work has provided important foundations in contract analysis, graph-based modeling, and anomaly detection. However, three gaps remain for early rug-pull warning in meme tokens: limited robust modeling under weak supervision and noisy labels, insufficient systematic validation of the chain from pattern profiling to warning scoring, and inadequate reporting of deployment-oriented indicators such as lead time and error boundaries.

To address these challenges, this paper focuses on early rug-pull warning for BSC meme tokens and proposes an end-to-end pipeline encompassing multi-granularity wash-trading pattern profiling, risk-feature construction, warning modeling, and error interpretation. The core idea is to convert Self, Matched, and Circular wash-trading behaviors into computable features, and then to use token-level risk vectors to drive supervised warning scores. Compared with methods relying on single indicators or static rules, the proposed approach emphasizes structured behavioral representation and reproducible implementation, aiming to improve both risk-ranking capability and interpretability. Under the current setting, Random Forest consistently outperforms Logistic Regression on key metrics ($\text{AUC}=0.9098$, $\text{PR-AUC}=0.9185$, $F_1=0.7429$) and provides usable warning windows for part of the samples (mean Lead Time~(v1) = 3.8133 hours).

The main contributions of this work are as follows. First, a multi-granularity wash-trading warning framework is established for BSC meme tokens, building a complete technical chain from on-chain data processing and pattern-feature extraction to warning modeling and error diagnosis, thereby enabling operational early identification of rug-pull risk. Second, a reproducible token-level behavioral feature system is designed comprising 12 features spanning transaction, address, temporal, and flow dimensions, and it is verified through main and ablation experiments that trade-level and address-level features are key contributors to risk ranking. Third, beyond standard classification metrics, Lead Time, FP/FN error profiles, and case timelines are reported, showing that the current model is better positioned as a high-precision screener and outlining a path toward higher-recall warning. Fourth, the framework preserves extensible interfaces for event-level timestamp labeling, time-split validation, and richer contract-mechanism features, providing a foundation for stronger statistical robustness and external validity.

The rest of this paper is organized as follows. Section~\ref{sec:method} presents the method and implementation details. Section~\ref{sec:results} reports experimental results together with discussion. Section~\ref{sec:conclusion} concludes the paper and outlines future directions.

\section{Method}\label{sec:method}

\subsection{Overview}

\begin{figure}[htbp]\centering\includegraphics[width=0.95\textwidth]{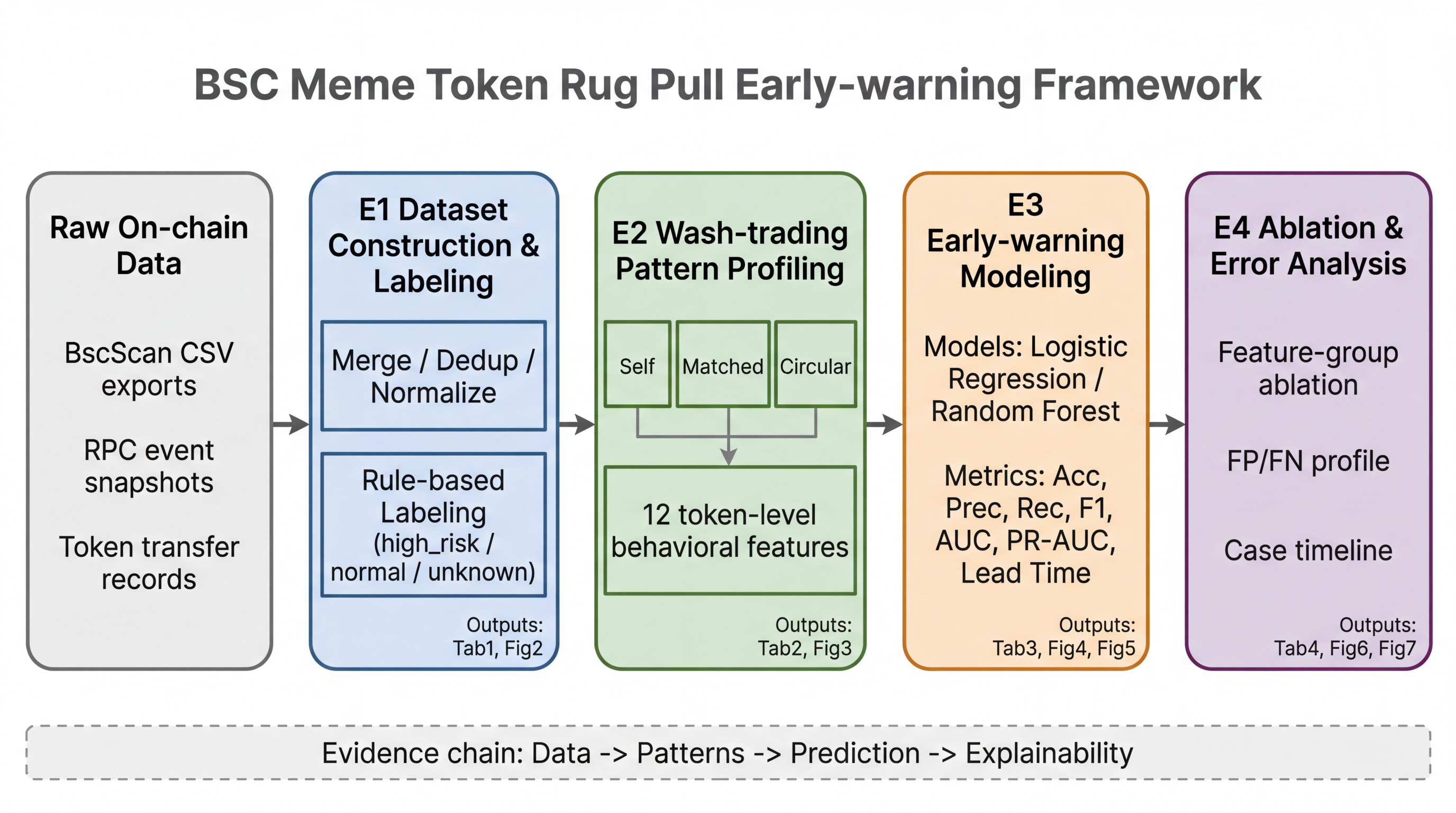}\caption{BSC Meme Token Rug Pull Early-warning Framework. The overall pipeline contains four stages (E1--E4): data construction and labeling, wash-trading pattern profiling, early-warning modeling, and ablation/error analysis.}\label{fig:framework}\end{figure}

An end-to-end framework is proposed for early rug-pull warning of BSC meme tokens. The overall pipeline, illustrated in Figure~\ref{fig:framework}, consists of four core stages: dataset construction and labeling~(E1), wash-trading pattern feature construction~(E2), early-warning modeling~(E3), and ablation/error analysis~(E4). The objective is to obtain interpretable risk-ranking capability and deployable warning signals under weak supervision and limited anomaly samples, using computable and reproducible on-chain behavioral features.

Unlike methods that rely on a single indicator or static rules, the proposed approach follows a three-layer pipeline: behavioral patterns, feature vectors, and model discrimination. Transaction behaviors are first mapped into quantifiable patterns, then aggregated into token-level risk features, and finally translated into warning decisions by supervised models. This design balances engineering feasibility and interpretability.

\subsection{Data Collection, Cleaning, and Labeling}

The raw data consist of BEP-20 token transfer events on BSC. To ensure cross-token comparability, up to the earliest 5,000 valid records are retained per token. This fixed-cap, unified-window strategy reduces direct bias from sequence-length differences. The current version contains 7 tokens and 33,242 records, as summarized in Table~\ref{tab:dataset}.

\begin{table}[htbp]\centering\caption{Summary statistics of the curated BSC meme-token dataset after preprocessing and deduplication.}\label{tab:dataset}\begin{tabular}{lr}\toprule Item & Value \\\midrule Number of tokens & 7 \\Total transfer records & 33,242 \\Per-token cap (records) & 5,000 \\Minimum records per token & 3,256 \\Median records per token & 5,000 \\Maximum records per token & 5,000 \\\bottomrule\end{tabular}\end{table}

\begin{table}[htbp]\centering\caption{Token-level record counts after preprocessing.}\label{tab:tokens}\begin{tabular}{lr}\toprule Token Address & Records \\\midrule\texttt{0x595d\ldots9cc} & 4,994 \\\texttt{0x4fa7\ldots3bf} & 3,256 \\\texttt{0x5440\ldots1b0} & 5,000 \\\texttt{0xe6df\ldots528} & 5,000 \\\texttt{0x8d0d\ldots0d} & 4,992 \\\texttt{0xe0a2\ldots2e} & 5,000 \\\texttt{0xcae1\ldots9d} & 5,000 \\\bottomrule\end{tabular}\end{table}

\begin{figure}[htbp]\centering\includegraphics[width=0.95\textwidth]{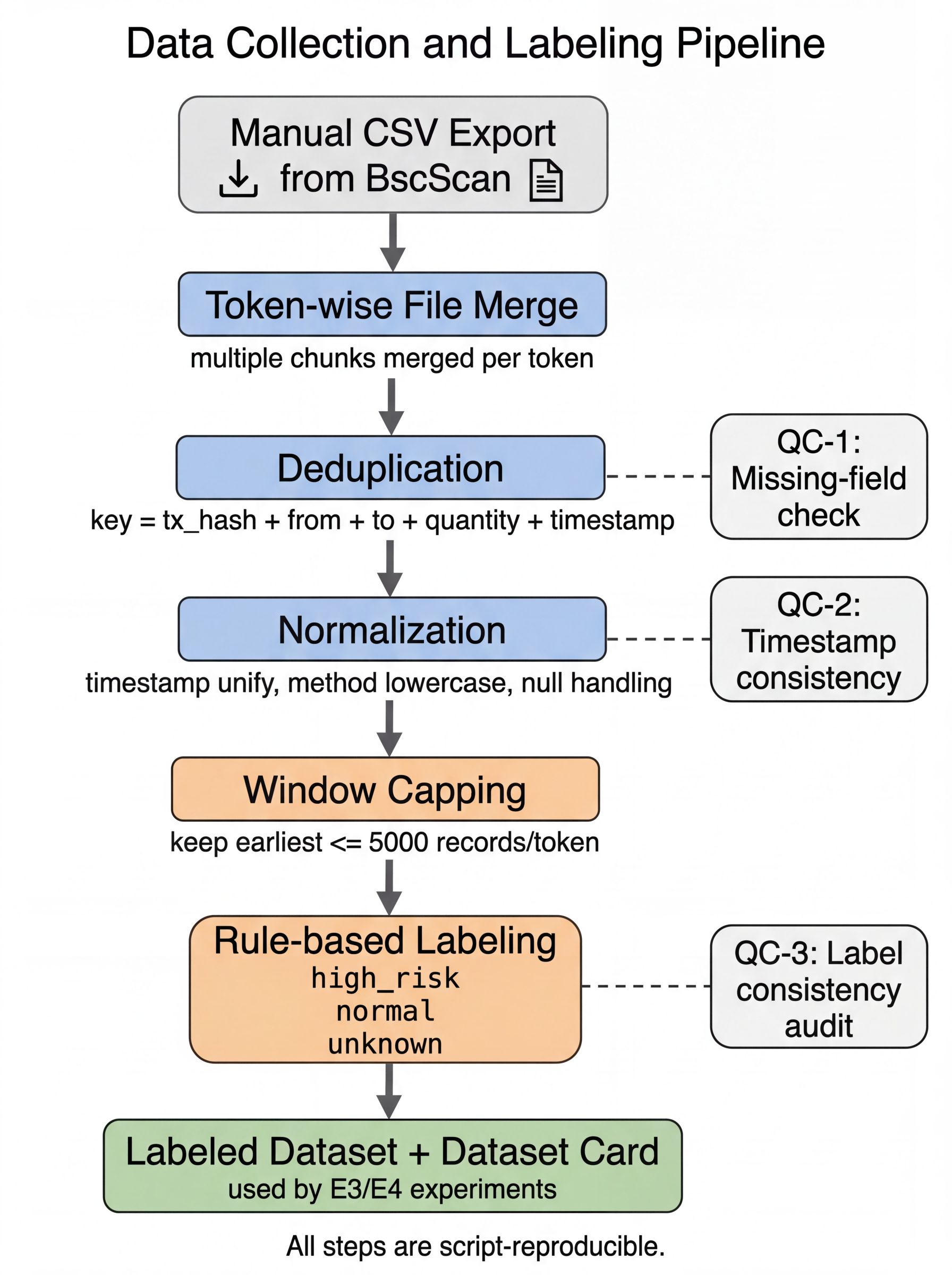}\caption{Data Collection and Labeling Pipeline. The workflow includes export, token-wise merging, deduplication, normalization, window capping, rule-based labeling, and quality-control checks.}\label{fig:pipeline}\end{figure}

The preprocessing pipeline, illustrated in Figure~\ref{fig:pipeline}, proceeds through five sequential steps. First, chunk merging combines multiple exported files by token address. Second, deduplication removes duplicated entries using the composite key ($\mathit{tx\_hash}$, $\mathit{from}$, $\mathit{to}$, $\mathit{quantity}$, $\mathit{timestamp}$). Third, normalization unifies timestamp format, field naming, and null handling. Fourth, window capping retains no more than the earliest 5,000 records per token. Fifth, consistency checks are run for missing fields, temporal order, and label auditing. All steps are script-based to ensure reproducibility and traceability.

Given current data constraints, a weak-supervision fallback is used for labeling: samples are categorized into \textit{high\_risk} and \textit{non\_high\_risk}, with \textit{unknown} temporarily merged into \textit{non\_high\_risk}. This setup is intended for validating the warning pipeline and maintains compatibility with future event-level upgrades (Lead Time~v2).

\subsection{Multi-granularity Wash-trading Feature Construction}

\begin{figure}[htbp]\centering\includegraphics[width=0.95\textwidth]{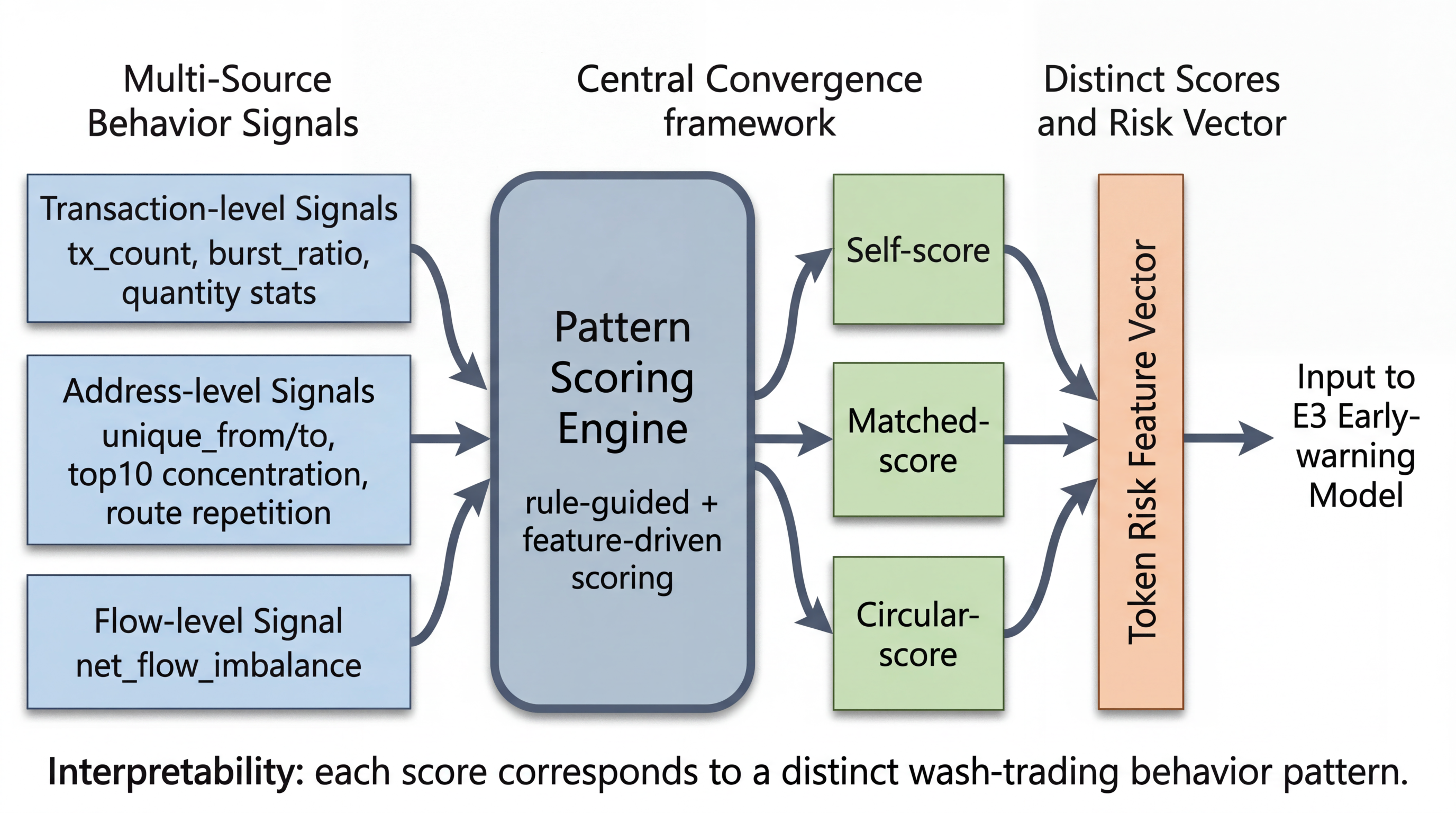}\caption{Wash-trading Pattern Profiling and Risk Vector Construction. Transaction-, address-, and flow-level signals are scored into Self/Matched/Circular components and then aggregated into token-level risk vectors.}\label{fig:profiling}\end{figure}

Three representative wash-trading patterns are modeled. The Self pattern captures self-looping or highly self-associated address behaviors. The Matched pattern identifies frequent paired transfers between address pairs. The Circular pattern detects multi-address cyclic fund transfers. As shown in Figure~\ref{fig:profiling}, transaction-, address-, and flow-level signals are fed into a scoring engine, which outputs three pattern scores and aggregates them into a token-level risk feature vector.

A total of 12 token-level behavioral features are constructed, as defined in Table~\ref{tab:features}. These features jointly describe activity intensity, structural asymmetry, transfer-size volatility, temporal burstiness, and suspicious route reuse, providing a unified representation for model learning.

\begin{table}[htbp]\centering\caption{Definition of token-level behavioral features used in early rug-pull warning.}\label{tab:features}\small\resizebox{\textwidth}{!}{\begin{tabular}{p{3.0cm}p{2.0cm}p{4.5cm}p{4.5cm}}\toprule Feature & Category & Definition & Interpretation \\\midrule $\mathit{tx\_count}$ & Activity & Number of transfer records & Overall trading activity \\$\mathit{unique\_from}$ & Addr.\ diversity & Unique sender addresses & Outgoing-side participation \\$\mathit{unique\_to}$ & Addr.\ diversity & Unique receiver addresses & Incoming-side participation \\$\mathit{from\_to\_ratio}$ & Struct.\ balance & $\mathit{unique\_from}/\mathit{unique\_to}$ & Sender/receiver asymmetry \\$\mathit{avg\_quantity}$ & Amount stats & Mean transfer quantity & Average transfer size \\$\mathit{median\_quantity}$ & Amount stats & Median transfer quantity & Typical transfer size \\$\mathit{std\_quantity}$ & Amount stats & Std.\ dev.\ of quantity & Transfer-size volatility \\$\mathit{top10\_addr\_ratio}$ & Concentration & Top-10 address share & Small-set dominance \\$\mathit{burst\_ratio}$ & Burstiness & Peak/median minute count & Burst-like trading \\$\mathit{net\_flow\_imbalance}$ & Flow imbalance & $|in-out|/(in+out)$ & Inflow/outflow imbalance \\$\mathit{route\_repeat\_ratio}$ & Route repetition & Repeated $(from,to)$/$\mathit{tx\_count}$ & Possible wash loops \\$\mathit{active\_minutes}$ & Lifespan & Active minute buckets & Temporal activity span \\\bottomrule\end{tabular}}\end{table}

\subsection{Early-warning Modeling and Inference}

\begin{figure}[htbp]\centering\includegraphics[width=0.95\textwidth]{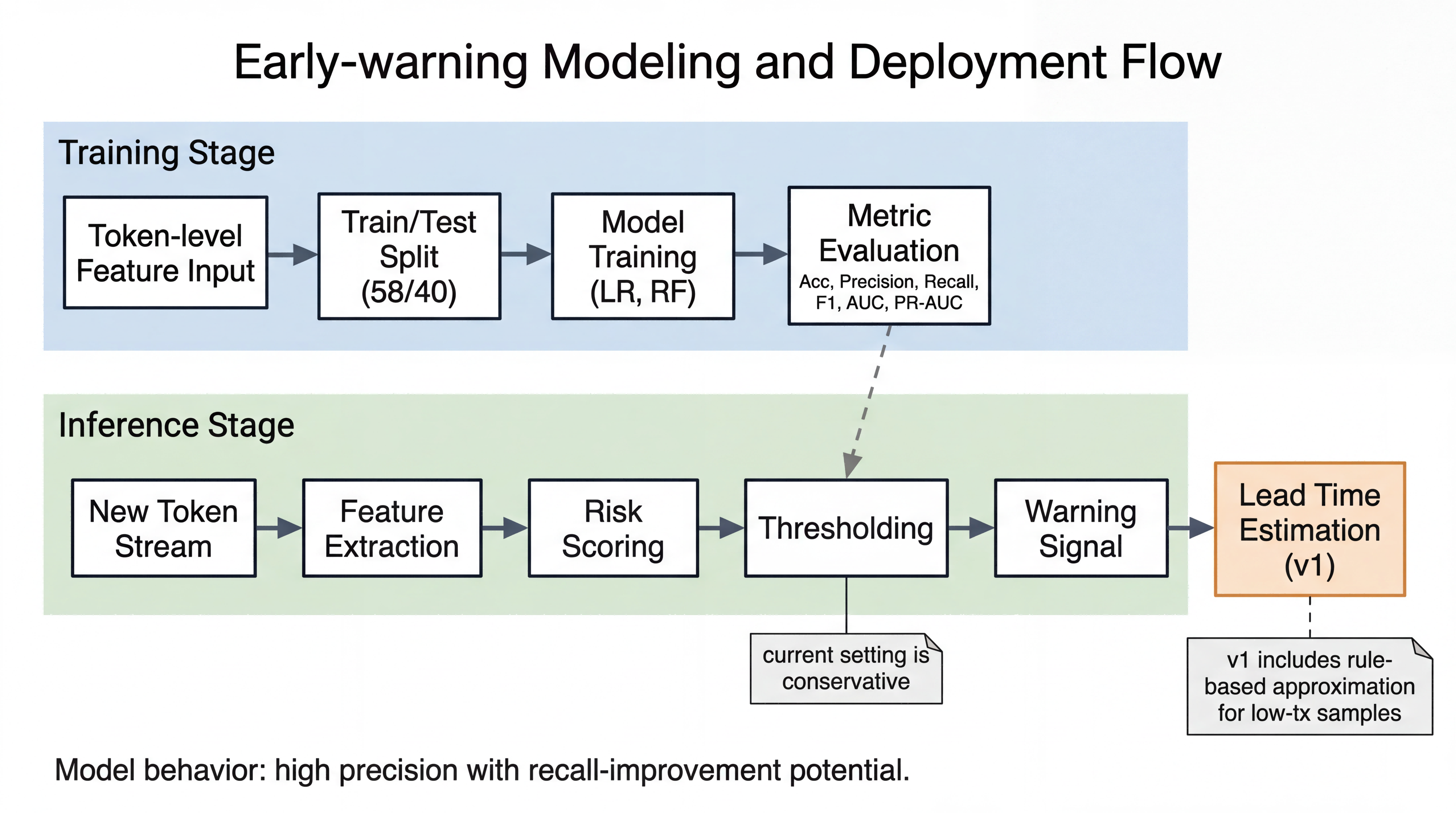}\caption{Early-warning Modeling and Deployment Flow. The upper lane shows model training and evaluation. The lower lane shows inference and alert triggering, with lead-time estimation as deployment output.}\label{fig:modeling}\end{figure}

Given a token-level feature vector $\mathbf{x}_i\in\mathbb{R}^{12}$, the model outputs a risk probability $\hat{y}_i\in[0,1]$. A warning is triggered when $\hat{y}_i>\tau$, where $\tau$ is a decision threshold. The training stage solves a binary classification task, while the inference stage outputs a risk score, a warning flag, and a lead-time estimate. The full train--inference loop is illustrated in Figure~\ref{fig:modeling}.

Logistic Regression and Random Forest~\cite{breiman2001random} are employed as the baseline and main model, respectively. Evaluation metrics include Accuracy, Precision, Recall, $F_1$, AUC, and PR-AUC. Lead Time (hours) is additionally reported to quantify warning timeliness. PR-AUC is treated as a key ranking metric due to class imbalance~\cite{chawla2002smote}. In the current version, Lead Time is defined as:
\begin{equation}
\text{LeadTime}=t_{\text{rugpull}}-t_{\text{warning}}
\end{equation}
where $t_{\text{warning}}$ denotes the first timestamp at which the predicted score exceeds the threshold $\tau$. For low-transaction samples (those with $\mathit{tx\_count}<2{,}000$), $t_{\text{rugpull}}$ is approximated in v1 by the last observed transaction timestamp. This assumption is treated as a bounded interpretation in subsequent analysis.

\subsection{Ablation and Error-analysis Design}

To quantify the marginal contribution of each feature group, three ablation settings are designed: removing trade-level features, address-level features, and contract-level features (currently a single feature), respectively. Each setting is compared against the full model via $\Delta\text{PR-AUC}$. In parallel, FP/FN distributions and case timelines (Figure~\ref{fig:timeline}) are analyzed to diagnose boundary behavior and deployment risk.

\subsection{Reproducibility and Implementation Details}

The entire pipeline is designed for full reproducibility, from raw CSV to token-level features to model outputs. All data-processing and feature-extraction scripts are provided as executable code. Main metrics, ablation results, and case timelines are saved as structured artifacts with explicit binding to the table and figure identifiers in this paper. It should be noted that the weak-supervision labels and Lead Time~(v1) approximation are phase-level validation settings and are not intended as final deployment claims.

\section{Results and Discussion}\label{sec:results}

\subsection{Dataset Construction and Pattern-feature Results}

A BSC-based dataset for early rug-pull warning was constructed following the procedure described in Section~\ref{sec:method}. After merging, deduplication, and normalization, the final dataset contains 7 tokens and 33,242 records (Table~\ref{tab:dataset}). Per-token record counts range from 3,256 to 5,000, with a median of 5,000, indicating a practical balance between multi-token coverage and per-token temporal depth for method validation. The complete token-level record counts are reported in Table~\ref{tab:tokens}, and the data processing workflow is illustrated in Figure~\ref{fig:pipeline}.

Based on the three wash-trading hypotheses (Self, Matched, and Circular), 12 token-level behavioral features were constructed (Table~\ref{tab:features}), covering activity intensity, address structure, route repetition, flow imbalance, and temporal burstiness. The profiling and feature aggregation pipeline is shown in Figure~\ref{fig:profiling}. This feature set provides a unified input interface for downstream warning models and supports feature-group ablation in subsequent analysis.

\subsection{Main Performance Comparison}

\begin{figure}[htbp]\centering\includegraphics[width=0.95\textwidth]{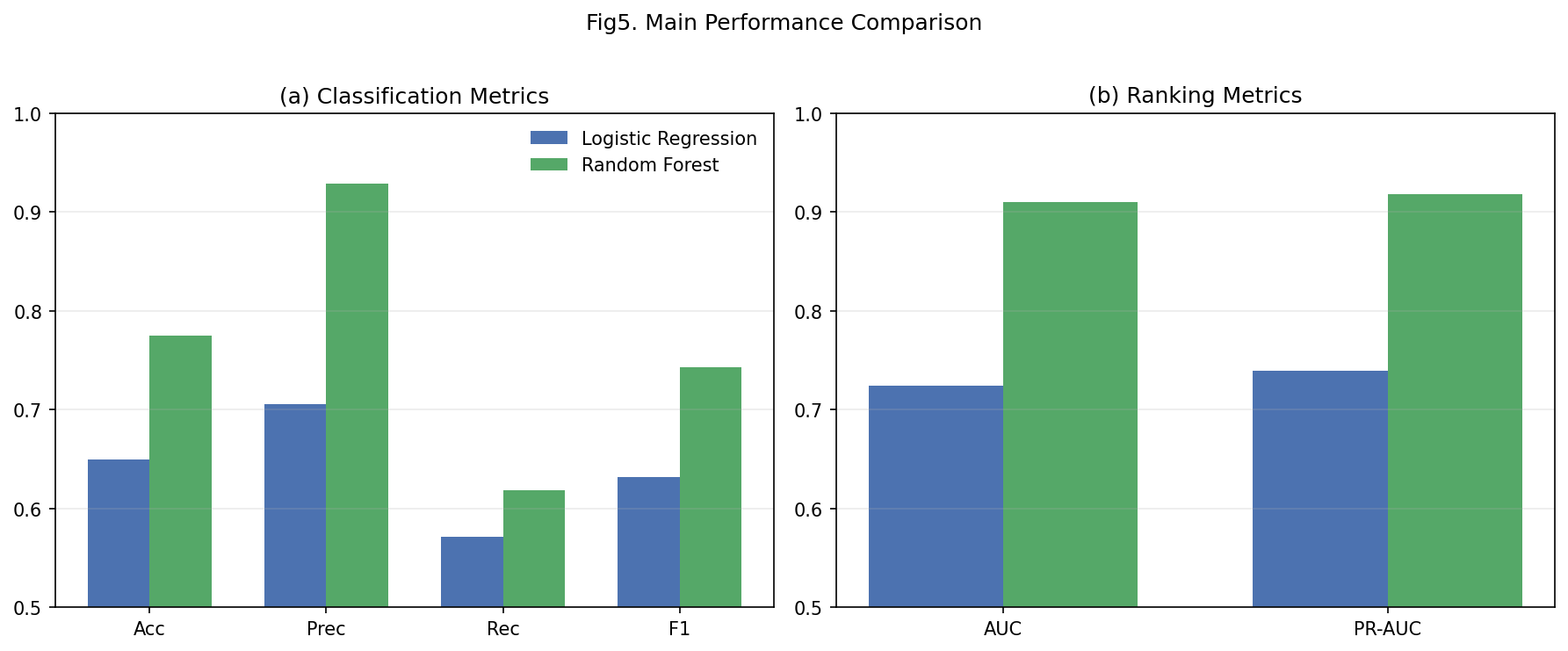}\caption{Main Performance Comparison between Logistic Regression and Random Forest. (a)~Classification metrics. (b)~Ranking metrics.}\label{fig:comparison}\end{figure}

Under the current split setting, Random Forest outperformed Logistic Regression on all core metrics, as reported in Table~\ref{tab:performance} and visualized in Figure~\ref{fig:comparison}. Specifically, Random Forest achieved $\text{Accuracy}=0.7750$, $\text{Precision}=0.9286$, $\text{Recall}=0.6190$, $F_1=0.7429$, $\text{AUC}=0.9098$, and $\text{PR-AUC}=0.9185$, whereas Logistic Regression obtained 0.6500, 0.7059, 0.5714, 0.6316, 0.7243, and 0.7397, respectively. These results indicate that, under the current weak-supervision setup, the non-linear model captures on-chain behavioral risk patterns more effectively.

\begin{table}[htbp]\centering\caption{Main early-warning performance comparison. Best values are boldfaced.}\label{tab:performance}\resizebox{\textwidth}{!}{\begin{tabular}{lccccccc}\toprule Model & Accuracy & Precision & Recall & $F_1$ & AUC & PR-AUC & Lead Time (h) \\\midrule Logistic Regression & 0.6500 & 0.7059 & 0.5714 & 0.6316 & 0.7243 & 0.7397 & 3.8133 \\\textbf{Random Forest} & \textbf{0.7750} & \textbf{0.9286} & \textbf{0.6190} & \textbf{0.7429} & \textbf{0.9098} & \textbf{0.9185} & 3.8133 \\\bottomrule\end{tabular}}\vspace{4pt}\\\raggedright\footnotesize\textit{Note.} Lead Time~(v1): $n=25$, mean$\,=3.8133$\,h, median$\,=1.0331$\,h, min$\,=0.2442$\,h, max$\,=54.4683$\,h.\end{table}

This performance gap indicates that non-linear models are better suited to capture the complex relationship between on-chain behavioral signals and rug-pull risk labels in this dataset. The value of this finding is not a claim of universally high performance, but rather empirical support for the operational chain from wash-trading pattern features to early-warning discrimination. In practical terms, the high precision (0.9286) implies a reduced false-alarm burden for downstream triage, while the recall level (0.6190) still leaves non-negligible missed-risk space that warrants future improvement.

\begin{figure}[htbp]\centering\includegraphics[width=0.85\textwidth]{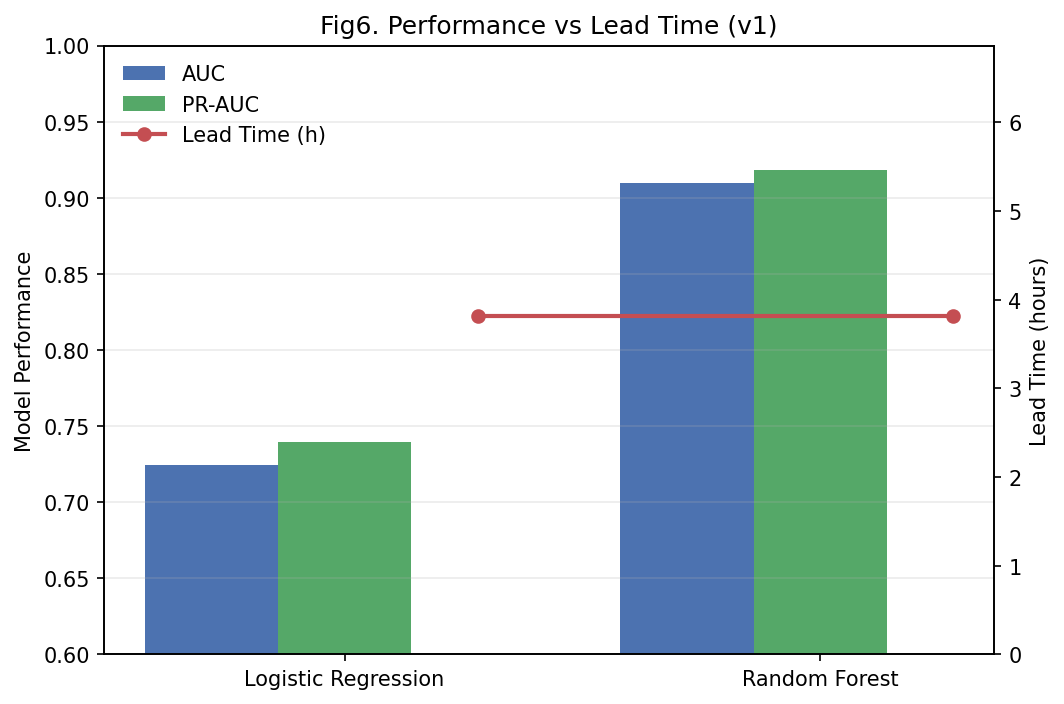}\caption{Performance vs.\ Lead Time Trade-off~(v1). AUC and PR-AUC are shown together with lead time.}\label{fig:leadtime}\end{figure}

Lead Time~(v1) reached a mean of 3.8133 hours ($n=25$) and a median of 1.0331 hours, as visualized in Figure~\ref{fig:leadtime}. This result suggests the existence of exploitable warning windows for a portion of the samples, although the lead-time distribution remains highly dispersed.

\subsection{Ablation Analysis and Feature-group Contributions}

Feature-group ablation results are reported in Table~\ref{tab:ablation}. Removing trade-level features reduced PR-AUC from 0.9185 to 0.7342 ($\Delta=-0.1843$), the largest drop among all settings, indicating that trade-level signals are the primary performance driver. Removing address-level features reduced PR-AUC to 0.8612 ($\Delta=-0.0573$), showing a stable complementary contribution. Removing the current contract-level feature yielded a slight PR-AUC increase to 0.9262 ($\Delta=+0.0077$), suggesting limited discriminative value of this feature group in the current version.

\begin{table}[htbp]\centering\caption{Feature-group ablation of the Random Forest model. $\Delta$PR-AUC is computed against the full model.}\label{tab:ablation}\begin{tabular}{lccccccr}\toprule Setting & \#Feat. & Prec. & Recall & $F_1$ & AUC & PR-AUC & $\Delta$PR-AUC \\\midrule Full model & 12 & 0.9286 & 0.6190 & 0.7429 & 0.9098 & 0.9185 & $+0.0000$ \\w/o trade & 6 & 0.7647 & 0.6190 & 0.6842 & 0.7168 & 0.7342 & $-0.1843$ \\w/o address & 7 & 0.7000 & 0.6667 & 0.6829 & 0.7920 & 0.8612 & $-0.0573$ \\w/o contract & 11 & 0.9231 & 0.5714 & 0.7059 & 0.9173 & 0.9262 & $+0.0077$ \\\bottomrule\end{tabular}\end{table}

The ablation analysis provides mechanism-level evidence for understanding the information hierarchy in the current model. The dominance of trade-level features ($\Delta\text{PR-AUC}=-0.1843$) suggests that trading behavior statistics constitute the most informative signal source. Address-level features offer a stable but smaller gain ($\Delta=-0.0573$), indicating complementary value from participant-structure cues. By contrast, removing the current contract-level feature slightly increases PR-AUC ($\Delta=+0.0077$), implying that this feature group is not yet sufficiently informative in its present form. This pattern is consistent with the design logic illustrated in Figure~\ref{fig:profiling}: risk is primarily identified through observable behavioral dynamics, while other feature groups serve as boundary refinements.

\subsection{Error Profile and Case Timeline Analysis}

The error profile on the test split is reported in Table~\ref{tab:error}. The system produced only $\text{FP}=1$ false positive but $\text{FN}=8$ false negatives, reflecting a conservative thresholding behavior with low false alarms but elevated missed detections.

\begin{table}[htbp]\centering\caption{Error profile of the Random Forest model on the test split.}\label{tab:error}\begin{tabular}{lccp{5.5cm}}\toprule Split & FP & FN & Observation \\\midrule Test split & 1 & 8 & Conservative thresholding with limited recall \\\bottomrule\end{tabular}\end{table}

\begin{figure}[htbp]\centering\includegraphics[width=0.95\textwidth]{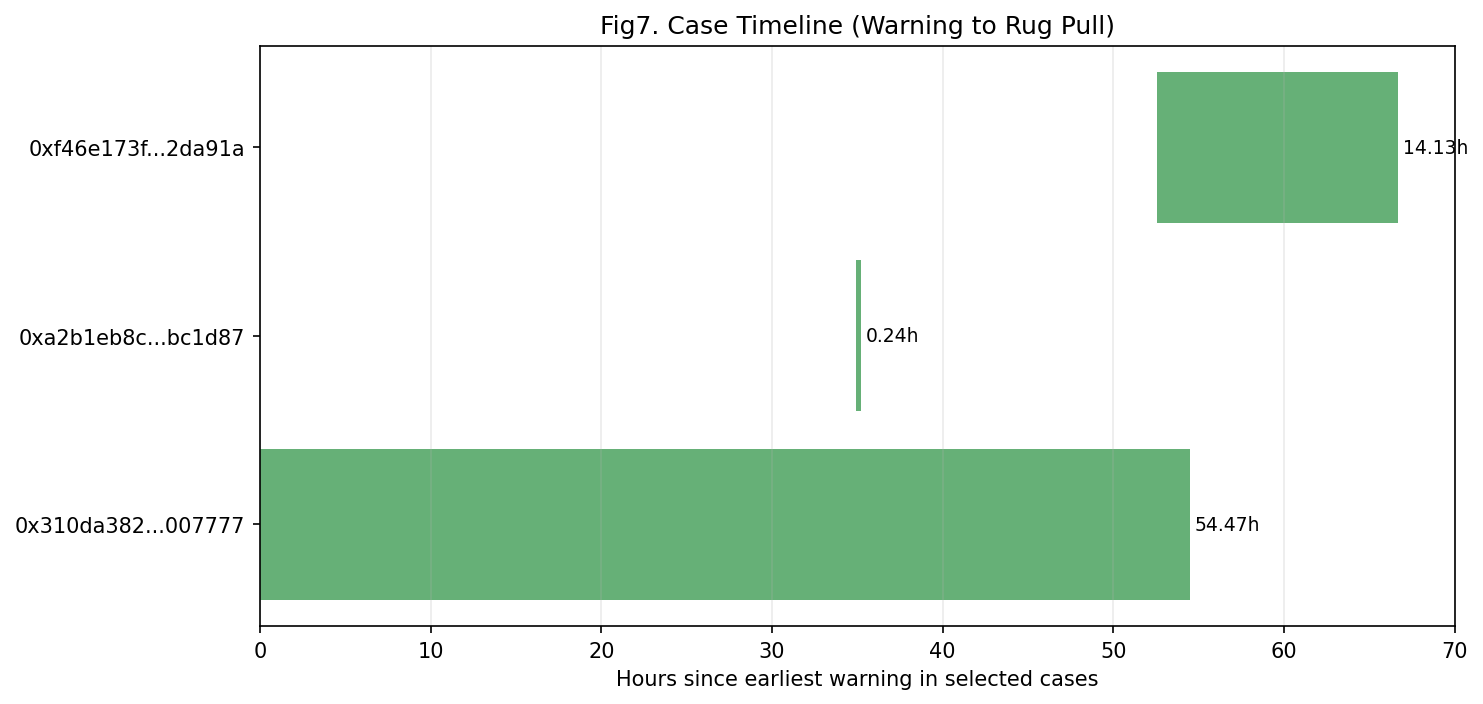}\caption{Case Timeline from Warning to Rug Pull. Representative cases visualize warning-to-event intervals and demonstrate heterogeneity in risk-evolution speed.}\label{fig:timeline}\end{figure}

Representative warning-to-event timelines are presented in Figure~\ref{fig:timeline}. Lead-time spans vary substantially across tokens, indicating heterogeneous risk-evolution dynamics.

From a deployment perspective, Lead Time~(v1) indicates preliminary operational value: a mean lead time of 3.8133 hours and a median lead time of 1.0331 hours suggest that actionable warning windows exist for a subset of tokens (Figure~\ref{fig:leadtime}). However, Figure~\ref{fig:timeline} shows substantial variation across case timelines, implying heterogeneous risk-evolution speeds. Combined with the error profile ($\text{FP}=1$, $\text{FN}=8$, Table~\ref{tab:error}), the current system is better characterized as a high-precision screener rather than a high-recall alarm engine. It is therefore more suitable for risk prioritization and human-in-the-loop triage than as a standalone automatic decision maker.

\subsection{Limitations}

Several limitations of the current study should be acknowledged. First, Lead Time~(v1) relies on a rule-based approximation for low-transaction samples, which is weaker than event-level timestamp evidence. Second, merging unknown samples into the negative class introduces label noise and may elevate false negatives. Third, confidence intervals, repeated-seed stability checks, and time-split external validation are not yet reported, limiting statistical robustness. Fourth, contract-level feature coverage is still narrow and does not fully capture mechanism-level risk factors. Future work should prioritize event-level $t_{\text{warning}}/t_{\text{rugpull}}$ annotation and Lead Time~v2, bootstrap confidence intervals and resampling-based stability reporting, train-on-past/test-on-future validation for stronger external validity, and richer contract-risk features such as privilege controls, tax mechanisms, and blacklist behaviors to improve recall and generalization.

\section{Conclusion}\label{sec:conclusion}

This study addresses early rug-pull warning for BSC meme tokens and validates an end-to-end framework covering on-chain data processing, wash-trading pattern profiling, risk modeling, and error interpretation. Centered on multi-granularity behavioral patterns, the framework unifies transaction-, address-, and flow-level signals into token-level risk features, and outputs warning scores through supervised models.

Under the current setting, Random Forest consistently outperforms Logistic Regression on key metrics ($\text{AUC}=0.9098$, $\text{PR-AUC}=0.9185$, $F_1=0.7429$), indicating a clear advantage of non-linear modeling for on-chain risk ranking. Ablation results further show that trade-level features are the dominant contributor, while address-level features provide stable complementary value and contract-level features still require enhancement. Lead Time~(v1) (mean 3.8133 hours), together with case timelines, suggests actionable warning windows for a subset of samples.

The contributions of this work are threefold. First, an executable, traceable, and reproducible rug-pull warning pipeline is provided that bridges the gap between wash-trading pattern detection and operational early-warning scoring. Second, empirical evidence is presented showing that multi-granularity wash-trading features remain effective under weak supervision, with trade-level and address-level feature groups driving the majority of risk-ranking performance. Third, deployment-oriented evidence is offered beyond standard classification metrics, including lead-time estimation and error-bound analysis, which informs the practical positioning of such systems as high-precision screeners suited for human-in-the-loop triage.

These conclusions are bounded by the current dataset scale, labeling strategy, and the approximated Lead Time~(v1) definition. Future work will focus on event-level timestamp annotation and Lead Time~v2, time-split external validation, bootstrap confidence intervals, and richer contract-mechanism features to improve recall, statistical robustness, and cross-scenario generalization.

\section*{Acknowledgments}
This research was supported by the Doctoral Start-up Fund of Baoshan University. The authors gratefully acknowledge the financial support provided by Baoshan University for this work.

\section*{Author Contributions}
\textbf{Dingding Cao:} Conceptualization, Methodology, Software, Formal Analysis, Writing -- Original Draft. \textbf{Bianbian Jiao:} Validation, Writing -- Review \& Editing. \textbf{Jingzong Yang:} Data Curation, Investigation. \textbf{Yujing Zhong:} Visualization, Validation. \textbf{Wei Yang:} Supervision, Funding Acquisition, Writing -- Review \& Editing.

\section*{Conflict of Interest}
The authors declare that there are no conflicts of interest regarding the publication of this paper.

\section*{Data Availability Statement}
The on-chain transaction data used in this study were collected from publicly available BEP-20 token transfer records on BNB Smart Chain (BSC). The processed dataset and feature-extraction scripts are available from the corresponding author upon reasonable request.

\bibliography{references}

\end{document}